# Estimating related words computationally using language model from the Mahabharata - an Indian epic

Vrunda Gadesha
*vrunda.g@ahduni.edu.in*

Keyur D. Joshi
*keyur.joshi@ahduni.edu.in*

Shefali Naik
*shefali.naik@ahduni.edu.in*

*Abstract*—'Mahabharata' is the most popular among many Indian pieces of literature referred to in many domains for completely different purposes. This text itself is having various dimension and aspects which is useful for the human being in their personal life and professional life. This Indian Epic is originally written in the Sanskrit Language. Now in the era of Natural Language Processing, Artificial Intelligence, Machine Learning, and Human-Computer interaction this text can be processed according to the domain requirement. It is interesting to process this text and get useful insights from Mahabharata. The limitation of the humans while analyzing Mahabharata is that they always have a sentiment aspect towards the story narrated by the author. Apart from that, the human cannot memorize statistical or computational details, like which two words are frequently coming in one sentence? What is the average length of the sentences across the whole literature? Which word is the most popular word across the text, what are the lemmas of the words used across the sentences? Thus, in this paper, we propose an NLP pipeline to get some statistical and computational insights along with the most relevant word searching method from the largest epic 'Mahabharata'. We stacked the different text-processing approaches to articulate the best results which can be further used in the various domain where Mahabharata needs to be referred.

## I. Introduction

Natural Language processing is the cutting-edge technology equipped with efficient tools and techniques to deal with unstructured text data. Using NLP pipeline techniques, a large amount of text can be processed very quickly and accurately. The most important point of processing the fictional text using NLP is that the text will be analyzed without adding any sentiments to it. 'Mahabharata' is the story orally often narrated and recreated across the world in different forms. Thus, humans have sentiments attached to them by default. So, to get the computational details about Mahabharata we used the elements of the NLP pipeline to answer the following questions which do not have any sentiment aspect attached with it. 1) How rich is 'Mahabharata' in terms of words? 2) Does the sentence length of 'Mahabharata' distribute normally across the whole literature? Apart from this, we are addressing the problem that how can we find the *most related* word from such a large text without reading it. In this paper, we are approaching the NLP Pipeline followed by the language model which is searching for the most related words from the large text.

## II. Literature Review

Mahabharata is a treasure of life lessons. To make this treasure of life lessons understandable for common people, it is important to translate it into the local languages used by people in daily life. The first translation of Mahabharata was written in the Persian Language entitled 'Razmnameh' on the order of Mughal Emperor Akbar in the 18th Century Later on, this is followed by English, Hindi, and other regional Languages [1]. The literature is narrating the phenomena and story which is lived by more than 200 people [2] which has been redacted between 400BCE and 400CE [3]. we can see the glimpse of various events that occurred in past across India and even across the globe[4]. Among these chunks, the city Bishnupur in West Bengal, India is famous for its terra-cotta temples. These temple's walls are carved with terracotta panels describing various events from 'Mahabharata'. These images are captured and used as a 3D image dataset known as BHID (Bishnupur Heritage Image Dataset) for various computer vision applications. BHID is a dataset containing a total of 4233 images which is in the public domain and is considered a central resource for research related to digital heritage [5]. The story of Mahabharata is retold in various art forms like plays, short stories, paintings, poems such as 'Kiratarjunyam' to make people understand the right ethics that not to make difference between 'High-man' and 'Low-man' where 'Lord Shiva' himself described as 'Kirat' [6] and translated books in various Indian languages. Though the Orality affects the translation, according to paper [7] the translation of literature may be treated as an independent text because "A study of translation is a study of language" and Mahabharata is retold in various Indian languages which can be a free translation or a literal translation. Here the difference between free and literal translation comes up in the picture because of the orality. Between these all forms of art, a unique art called 'Wayang (leather puppets)' is famous for recreating the Mahabharata story in Bali - Indonesia. Sudiatmika (et al. 2021) and fellow researchers have classified the 'Mahabharata Events' presented in this art form. They used the R-CNN algorithm to achieve the recognition of events and the characters such as 'Wayang Arjuna' and 'Wayang Yudhistira' [4]. People are always interested in hearing or watching fictional or fantasy stories. Thus, stories inside Mahabharata are always being attractive for creative people. This Epic is even inspiring for the technologist to create various taxonomies for the fictional domain (TiFI) [8] and launch 'ENTYFI' – the first technique for typing entities in fictional text. This 5-steps technique is useful to generate supervise fiction typing, supervised real-world typing and unsupervised typing [9]. A large number of events and characters in the epic is also useful for Ontology (a knowledge representation structure). In the current scenario, the web resources are more explored for ontology enrichment rather than the question-answer-pair (QA-pair). Authors in paper [10] applied such QA-pair on the 'Mahabharata Domain' to convert them into potential triples (subject, predicate, and object) and identify the triples which are new, more precise, and related to the domain for ontology enrichment in literature. During an ACM conference on 'Data and Application Security and Privacy (CODASPY) in a panel session prof. Rakesh Varma compared the data security issues with the Mahabharata War. He mentioned that in this world of data we are facing an untold war where attackers are motivated and working more sophisticatedly rather than we are fragmented with our data [11]. Apart from the angles like literature, Security, Technology, Digital Heritage Research, text generation, or literature of translations, Mahabharata is referred for the analysis of the 'Ludo Game' played on an android device. The analysis of different nine games concluded that the face of the dice is not equally distributed. Thus, the dice is biased and the dice algorithm is designed in such a way to make the game closer for an exciting experience for the users. Authors in paper [12] took the context from the history as well that the ludo game is inspired from that game called 'Pachishi' which is similar to the game played in Mahabharata called 'Chaupar'. While looking at the various aspect of Mahabharata, excluding the psychological aspects is not possible. Authors in paper [13] has explored the evidence for the most fundamental metaphor used for the mind – "The Mind is a Container" in Indian Epic 'Mahabharata' and 'Ramayana' plus the Greek Epic poem 'Homer and Hesiod' to traverse the cognitive phenomena in the epic literature. This study provides many uncommon aspects of our mental life. The description of the concept of the mind container is elaborated on the base of the epics by

(a) Ascription, Location, the content of the mind container, (b) Scope of mind container concerning consciousness and memory, (c) control over the content and (d) functions of the mind container.

Mahabharata Wiki Article is featured in the 100 most viewed Wikipedia article list. It is easy to give the context of the literature to people who belong to different domains. Thus, it is important to have a computational, analytical, and sentimental analysis of the text to get meaningful insights [14] In [15], the authors have derived interesting insights from the English translation of Mahabharata [2] by applying Pre-processing, POS tagging, Co-occurrence analysis, sentiment analysis of text and characters, and emotional analysis. The Insights which are given about the character and phenomena are versatile enough to use in different domains. According to [16] paper, The important characters of the epic Arjuna and Bheema had a common struggle and they trust each other abilities more intensely, this is also derived by [15] in the sentimental analysis across the text that "Arjuna and Bheema faced more negativity around them". In paper [16], the author brings the concept of considering the human values while designing the AI Agents. The similarity between humans and AI agents is positively correlated to the trust factor. Thus, inspired from the story of 'Challenging the powerful kings like Jarasandha and Chitrasena by arjuna and Bheema' can help AI-Agent developers to involve Value similarity in the outline of AI-Agent development. Apart from the technology development, the treatise has relevance to the modern society and is helpful to derive management lessons such as Strategic Management, Creation and relation with powerful friends and Allies, Effective Leadership Style, Successful Team Building, Shared goal and Ownership of the Goal, Commitment to the Goal, Role Clarity, Understanding the ground realities and Empowering Women [17]. The most important part of this Epic is 'Bhagavad Gita' said during Bhishma Parva also gives lessons of intrapersonal skills like Self-development, sublimation/management of the physical dimensions, sublimation/management of the psychological dimensions, Deontology, desire management, anger management, mind management, Emotional Stability, Fear Management, self-motivation, Empathy, and social welfare [18]. This epic gives the zoom version of the art of concentration with the lifespan of Arjuna. Different events can lead us to derive the factors which can be considered for the concentration like Enthusiasm, Dedication, Aptitude, Emotional or Physical state, and Environment [19]. The Epic context is shaping the thinking of society over the centuries. And this is reflected in our modern literature for children and adults. The stories derived from the epic show the disability as a curse or sin, but the modern literature shows the positivity and power of the disability. It portrays the usefulness of disabled people to society. In the context of Mahabharata, the approach towards the disability may fall under the bucket of "Don'ts"[20].

## III. Methodology

This paper aims to carve the non-semantic, statistical, and computational insights along with finding the most relevant words from the largest Indian Epic 'Mahabharata'.Figure-1 shows the NLP pipeline, which is defined to get robust results on the text. During this experiment "The Mahabharata of Krishna Dwaipayana Vyasa – The English translation (1886-1889) by Kesri Mohan Ganguli" is used in '.EPUB' format as a dataset.

### A. .EPUB file conversation into data structure

The '.EPUB' – (electronic publication) format is a very popular format of the e-book in digital documentation. This format is not only useful to read e-books using multiple devices such as android/mac mobiles, tablets, laptops, or desktops but these files are also useful for text processing. The EPUB format is released as an archive file built on the XHTML method. The tag format of XHTML can be flattened into any data structure which is readable by machine language. Here the whole e-book is converted into a python list Data structure.

As shown in above figure 2, the 'Mahabharata' e-book is divided into sequential data structure. While conversation the 'New line' is converted into '\n' and page break is converted into '\0'. apart from this, we do have some unwanted elements such as comma (,) semicolon (;) and apostrophe 's' ('s).

### B. Text Cleaning

The Mahabharata story contains many punctuations which are important to understand the sentiments for humans, but it not useful for the machine. Text cleaning is addressing the problem to handle unwanted elements. Using python library 're' (Regular Expression) and 'string' the redundant elements such as comma (,) semicolon (;) and apostrophe 's' ('s) are removed from the whole text and the text is now stored a unit string. In the general case, full-stop (.) is also removed during the text cleaning of the data-set but this process required full-stop (.) while performing the next step of the pipeline called tokenization. the reason behind keeping the full stop is to define the end of the sentences. After tokenization, we can find the number of words occupied in each sentence which can be identified as a word distribution pattern.

### C. Tokenization

The concept of dividing the text document into small snippets is known as tokenization. The tokenization can be applied in two different ways on the text document: (a) Sentence tokenization and (b) word tokenization. These can generate a bunch of sentences, words, phrases, tokens, or symbols [21]. Usually, Tokenization is applied as a primary and conventional text-preprocessing step in an NLP pipeline.

In the text-preprocessing of the Mahabharata, we used the 'Natural Language toolkit – sent_tokenize()' method to divide the whole text into sentences. The whole Mahabharata is divided into 1,30,700 sentences with variable lengths. The length distribution is described in figure 4.

As shown in figure 4, most of the sentences have a length between 20 to 70 words. And very few sentences are having a length of less than 20 and some outliers do have higher lengths like the sentence on the 121306 index is having a length of 1850 words.The text is not only divided into chunks of sentences but also into unit words to add more granularity into text preprocessing. This is achieved using the technique called word tokenization. Here we used the 'Natural Language Toolkit – word-tokenize()' method which divides the whole Mahabharata text into 27,49,461 uni-grams (only one word).

### D. Text Normalization

The human written text includes Function Words and Content words. These text data, specifically the fictional text is a combination of all the grammatical ups and downs. Thus, these data do have high randomness. to reduce the randomness of the text and maintain the significant meaning of the text, the text normalization can be performed on the whole text.

On the Mahabharata text, we are applying two popular techniques Stemming and Lemmatization. These tasks are followed in the NLP pipeline to transform the fictional text into the standard form of the language. Both these tasks are followed by 'Removing Stop Words' on the text.

In Mahabharata text, many words do not have critical significance but are used with high frequency throughout the whole epic to form the correct grammar. these words are not useful to improve the performance of any language model and they will also take some computational time in the further analysis process. These words do not have any information in terms of sentiment analysis as well. So, it is advisable to remove stop words (words like a, an, the, are, have, etc. ) along with the text normalization tasks.

*1) Stemming with Stop Words:* One word having the same semantic meaning can be written in many formats with human language. Stemming is a technique that removes the affixes attached to the word and tries to bring out the stem word or root word from the text. Among popular stemming techniques like Lancaster stemmer, Porter Stemmer, and Snowball stemmer, we used Porter stemmer to get the root words of the whole Mahabharata text.

*2) Lemmatisation with stop words:* The process of Lemmatisation is designed with the same purpose which is addressed by stemming. Lemmatisation is also used to cutting down the words to their root word. However, in Lemmatisation, the inflection of the word is not just broken-off, but it uses the concept of lexical knowledge. Using this converts the words into base form. Thus, it holds the sentiments of the text more strongly. Here we used 'wordnetlemmatizer' to achieve this task.

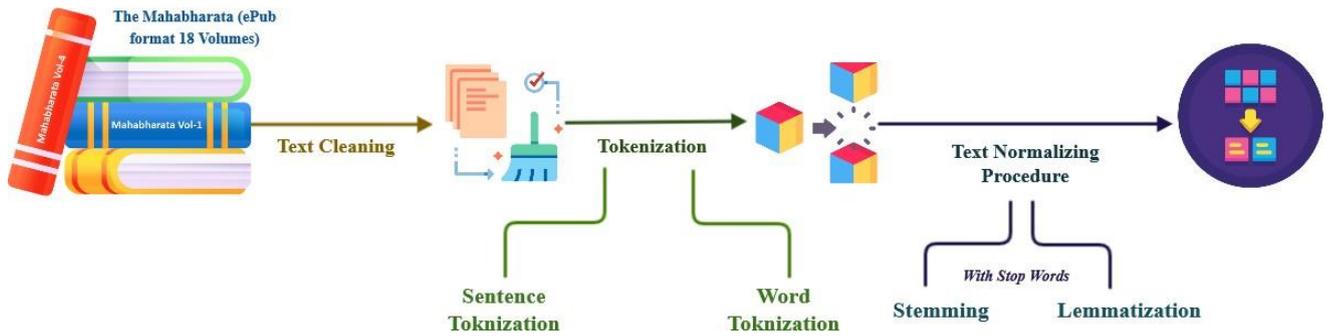

Fig. 1: Text pre-processing Pipeline on The Mahabharata

```
In [11]:  out = epub2text('mahabharata.epub')

In [12]:  out

Out[12]:  ['\n \n \n \n \n ',
           'Adi Parva \n \n \n Translators Preface \n Section 1 \n Section 2 \n P
          aushya Parva \n Pauloma Parva \n Astika Parva \n Adivansavatarana Parva
          \n Sambhava Parva \n Jatugriha Parva \n Hidimva-Vadha Parva \n Vaka-Vad
          ha Parva \n Chaitraratha Parva \n Swayamvara Parva \n Vaivahika Parva \
          n Viduragamana Parva \n Rajya-Labha Parva \n Arjuna-Vanavasa Parva \n S
          ubhadra-Harana Parva \n Haranaharana Parva \n Khandava-Daha Parva \n \n
          \n ',
           'Translators Preface \n The object of a translator should ever be to h
```

Fig. 2: The Mahabharata - in a list structure

```
In [21]:  mahabharata_str_cleaned

Out[21]:  '       adi parva     translators preface    section 1    section 2
          paushya parva     pauloma parva    astika parva    adivansavatarana parva
          sambhava parva    jatugriha parva    hidimva-vadha parva    vaka-vadha par
          va    chaitraratha parva    swayamvara parva    vaivahika parva    vidurga
          mana parva    rajya-labha parva    arjuna-vanavasa parva    subhadra-haran
          a parva    haranaharana parva    khandava-daha parva       translators p
          reface    the object of a translator should ever be to hold the mirror u
          pto his author. that being so   his chief duty is to represent so far as
          practicable the manner in which his author s ideas have been expressed
```

Fig. 3: The Mahabharata - after text cleaning stored as unit string.

The selection between stemming and Lemmatisation can be done based on the database on which the Language model is going to be built. The Mahabharata is a fictional text, and to extract features from this large epic, a strong sentimental hold on the text is required. Thus, based on the comparison of stemming and Lemmatisation we decide to build the language model on lemmatized text

### E. The Language Model

The second objective of this paper is to find similar words from the Mahabharata fictional text. So basically, we are targeting to implement a model which can process as illustrated in figure 6.

Here we have a large amount of fictional text which can be considered as unannotated data for training a model. Thus, according to [21] word2vec is well liked model to be applied on data which do not have any adulteration.Word2vac is a combination of two different algorithms applied together on corpus. These two algorithms are known as CBOW (Continuous Bag of Words) and Skip-Gram. This model is developed with three different layers. (a) Input Layer, (b) Single Hidden Layer and (c) Output Layer. The input layer is consisted with set of neurons which is having shape of the total number of words in the vocabulary. This vocabulary is specifically built according to the corpus. In this paper our corpus is the book "Mahabharata" and the vocabulary created

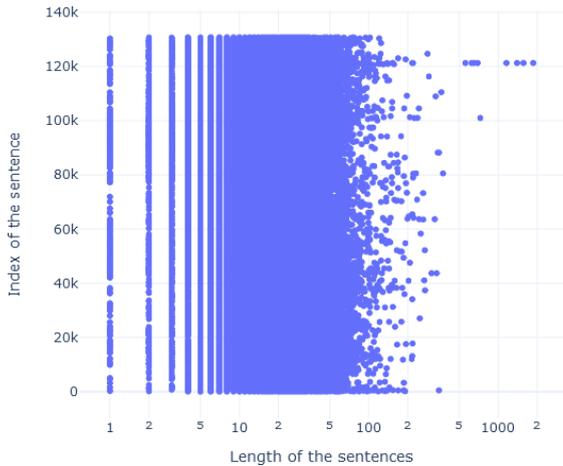

Fig. 4: Sentence Distribution of Mahabharata

(figure - 7) based on this text is containing 25794 words.

The magnitude of a single hidden layer is equal to the dimensionality of the result word vector. Here we trained a word2vec model to get the 100-dimension resultant vector. So, the size of the hidden layer is 100 dimensional. And the output layer is having the same magnitude as the input layer. Considering 'V' words in the vocabulary (where V=25794) and 'N' is the dimension of the resultant vector (where N=100). Thus, the connections from the input layer to the hidden layer can be constituted by the WI matrix having the shape of V × N. Here each row and column represents each word of vocabulary and the dimension of the resultant vector respectively. Likewise, the connections from the hidden layer to the output layer can be constituted by a WO matrix having the shape of N × V. Here each row and column represents the dimension of the resultant vector and each word of vocabulary respectively.

Considering the above sample corpus (figure 8), the vocabulary created based on this corpus can be represented as follows:

Vocabulary$_s$ = 'one' : 0, 'day' : 1, 'wait' : 2, 'upon' : 3, 'wrathful' : 4, 'ascetic' : 5, 'rigid' : 6, 'vow' : 7, 'durvasa' : 8, 'name' : 9, 'acquainted' : 10, 'truth' : 11, 'fully' : 12, 'conversant' : 13, 'mystery' : 14, 'religion' : 15, 'pritha' : 16, 'possible' : 17, 'care' : 18, 'gratified' : 19, 'rishi' : 20, 'soul' : 21, 'complete' : 22, 'control' : 23, 'holy' : 24, 'attention' : 25, 'bestowed' : 26, 'maiden' : 27, 'told' : 28, 'satisfied' : 29, 'fortunate' : 30, 'thee' : 31, '!' : 32

The sample corpus vocabulary has 33 words. This vocabulary is considering each unique word given in the sample corpus. So, there are 33 input neurons and 33 output neu-rons. We have 100 neurons in the hidden layer. Thus, our connections neurons between the input layer to the hidden layer can be represented as WI(33 ×100) and the connection neurons between the hidden layer to the output layer can be represented as WO(100 × 33). Now before we train the word2vec model these matrices are initialized with small random numbers.

Now looking at the corpus, if we want that word2vec model finds the relationship between the words "durvasa" and "vow"; the word "durvasa" is known as context, and "vow" is known as the target.

Now, these inputs can be multiplied with the randomly initialized WI(33 ×100) matrix tending towards the hidden layer, and then the output at the hidden layer will be multiplied with WO(100 × 33)matrix while tending towards the output layer. The target of this model is to compute probabilities for words at the output layer. This is achieved in word2vec as it implements the softmax function.

The idea behind using word2vec is, this model is used to represent the words by a vector of numbers. In our case, we provide the target word as input to the model. It will compute the cosign similarities between all other words available in the vocabulary and send it back as output with *top n* words.

## IV. Results

After Text-preprocessing we applied word2vec on the corpus with 25794 length vocabulary. we considered 'Hastinapur(location)', 'Arjuna (protagonist)', 'Gandiva (object)', 'Shakuni (protagonist)', 'dice (object)', 'Krishna (protagonist)' and 'Siva (as Character)' as target words. These words are selected based on the popularity of the protagonist, location, and object covered in Mahabharata. The vector representation of the word is illustrated in figure 9.

The sample target words with their similar words along with the cosign similarity between target word and context word is shown table-1. *(Here we are considering top five similar words)*

## V. Conclusion

In this paper, the NLP-based experiment on the Mahabharata is carried out on a basic level. We trained the word2vec model on the corpus to get the most similar words from the text itself. The reason behind selecting word2vec is to deal easily with the high-dimensional word vectors. In this paper, we can reach the basic aspects like uni-gram vocabulary, sentence distribution, 100-dimensional vector representation, and word similarities.

## VI. Future Scope

Though in the current scenario we do have similar models like 'Glove' and 'Fast Text' which we are targeting to apply and compare the results. The comparison of these models will bring a robust argument that which model is giving the best result on fictional text. Apart from the text-similarity, we are

```
# Original Sentence
sentences[470]

'the endeavours of duryodhana to engage yudhishthira again in the game; a
nd the exile of the defeated yudhishthira with his brothers.'

# Sentence after Stemming
sentences_stm[470]

'endeavour duryodhana engag yudhishthira game ; exil defeat yudhishthira
brother .'

# Sentence after lemmatisation
sentences_lemma[470]

'endeavour duryodhana engage yudhishthira game ; exile defeated yudhishth
ira brother .'
```

Fig. 5: Stemming vs Lemmatisation

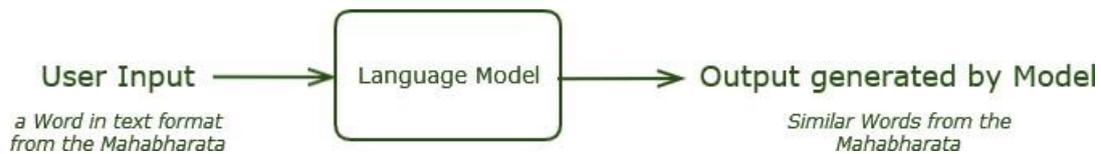

User Input → Language Model → Output generated by Model

*a Word in text format
from the Mahabharata*

*Similar Words from the
Mahabharata*

Fig. 6: Language Model Overview

| Index | Vocab Words | Index | Vocab Words |
|-------|-------------|-------|-------------|
| 4782 | behaves | 29391 | palala |
| 32282 | arthakaman | 14122 | diffusing |
| 25137 | anangangahara | 25456 | bhuyas |
| 14308 | bread | 15506 | pravarakarna |
| 2529 | fulfilled | 13356 | ananda |
| 9092 | overspreads | 24286 | urdhvavahu |
| 10737 | vacant | 19381 | buyest |
| 22088 | durvaranah | 5034 | secretion |
| 4901 | befallen | 8882 | brahma-weapon |
| 28502 | dipaka | 28471 | well-instructed |

Fig. 7: Vocabulary based on Mahabharata Corpus

```
sentences_lemma[3671]

'one day wait upon wrathful ascetic rigid vow durvasa name acquainted tru
th fully conversant mystery religion .'
```

```
sentences_lemma[3672]

'pritha possible care gratified wrathful rishi soul complete control .'
```

```
sentences_lemma[3673]

'holy one gratified attention bestowed maiden told satisfied fortunate on
e thee !'
```

Fig. 8: Sample Corpus from The Mahabharata

also targeting to perform noun extraction, word cloud for the various protagonist of Mahabharata, creating custom corpus, parts-of-speech-tagged word corpus, chunked phrase corpus, and create categorized text corpus of Mahabharata to make it open source for various experiments. Apart from these, we are targeting text classification on the Mahabharata.The major purpose to opt for these NLP techniques on Mahabharata is to understand and observe the behavior of the Mahabharata

```
# 100 dimension vector
vector

array([[ 0.17051627,  0.05652179, -0.04896212, -0.20656443,  0.12825224,
        -0.5205791 ,  0.12808476,  0.6689745 , -0.18518427, -0.4003636 ,
        -0.1872233 , -0.22397001, -0.42418396,  0.10391963,  0.46836236,
        -0.36834765, -0.01753333, -0.10635588, -0.10658871, -0.35558546,
         0.0699246 , -0.02912686, -0.04366722, -0.3427552 ,  0.37802848,
        -0.02670253,  0.11389722, -0.28978613,  0.12131397,  0.09461687,
         0.13793804,  0.2581571 ,  0.12898628, -0.12989827, -0.03690538,
         0.27640244,  0.38282153,  0.23045439, -0.45278904, -0.46427873,
        -0.07617377,  0.14015028,  0.15100801,  0.16213828,  0.24958597,
         0.32573652, -0.15923294,  0.13431995, -0.21606189,  0.22865067,
         0.26642472, -0.46106708, -0.1558434 ,  0.16029985, -0.10055732,
         0.10350485,  0.16122694,  0.16974118,  0.3864743 ,  0.489417  ,
         0.01043846,  0.68263197, -0.11287693, -0.2368424 , -0.04398678,
         0.12122558,  0.13474753,  0.07412283,  0.11234121,  0.3853903 ,
         0.18256441, -0.28511426,  0.20384145, -0.33893153, -0.33626717,
         0.1658625 ,  0.18308787, -0.5240451 , -0.3906647 ,  0.2834786 ,
        -0.16816811,  0.42461804, -0.08096772, -0.44269383,  0.19216105,
        -0.07874577,  0.23114307,  0.34356585,  0.04472911, -0.06908701,
         0.5040903 ,  0.08982632, -0.19615307,  0.18033302,  0.21443933,
        -0.01323832,  0.6281646 , -0.08676291, -0.06227214,  0.02094649]],
      dtype=float32)
```

Fig. 9: Vector Representation from Mahabharata Vocab

| context word | cosine similarity |
|---|---|
| **Sample target word: Hastinapur (Location)** | |
| sheltered | 0.8325892090797424 |
| vaivaswata | 0.8262888193130493 |
| Ansu | 0.8186947107315063 |
| coronet | 0.8173437714576721 |
| kurujangala | 0.8155612345510193 |
| **Sample target word: Arjuna (Protagonist)** | |
| partha | 0.8605594038963318 |
| Dhananjaya | 0.8489114046096802 |
| karna | 0.8209961896048279 |
| Vibhatsu | 0.7797690629959106 |
| phalguna | 0.7688585519790649 |
| **Sample target word: Gandiva (Object)** | |
| discus | 0.8680269112123718 |
| Wielder | 0.8528252840042114 |
| mace | 0.7967674732208252 |
| thunder-bolt | 0.7912233471870422 |
| trident | 0.7857180833816528 |
| **Sample target word: Shakuni (Protagonist)** | |
| shikhandi | 0.9232674241065979 |
| somadatta | 0.9199751615524292 |
| duhsasana | 0.9096741167630566 |
| vikarna | 0.9092869758605957 |
| satanika | 0.9051489233970642 |
| **Sample target word: dice (Object)** | |
| war | 0.7768427133560181 |
| suyodhana | 0.7763647437095642 |
| Match | 0.7677909135818481 |
| Wretched | 0.7589513659477234 |
| Gambling | 0.7584301829338074 |
| **Sample target word: Krishna (protagonist)** | |
| kesava | 0.8743714690208435 |
| janardana | 0.7793148756027222 |
| vasudeva | 0.7790760397911072 |
| Keshava | 0.768439769744873 |
| vibhatsu | 0.7160406708717346) |
| **Sample target word: siva (character)** | |
| Sthanu | 0.9016269445419312 |
| three-eyed | 0.900800083769989 |
| boon-giving | 0.9007714986801147 |
| isana | 0.8950022459030151 |
| skanda | 0.8801624774932861 |

TABLE I: Table - 1 cosine similarity between words in Mahabharata

protagonist. These observations can be matched with human behavior with the help of a specific questionnaire based on organizational behavior. This can provide a profile of a person and production capacity in his/her working environment. Thus, the results of this paper can be mapped with future research to identify the professional perspective of a human personality based on the Mahabharata.